\documentclass{new_tlp}

\usepackage{color}
\usepackage{comment}
\usepackage{hyperref}
\usepackage{amsmath}
\usepackage{xspace}
\usepackage{enumitem}
\usepackage{rotating}
\usepackage{multirow}

\def\AIA{${\cal AIA}$\xspace}
\def\oldTI{$old{\cal TI}$\xspace}
\def\newTI{$new{\cal TI}$\xspace}
\def\newoldTI{$new$+$old{\cal TI}$\xspace}
\def\newTIonly{$new{\cal TI}$-only\xspace}
\def\kb{${\cal KB}$\xspace}
\def\var#1{{\hbox{{\bf #1\/}}}}
\def\inst#1{{\hbox{{\tt #1\/}}}}
\def\sort#1{{\hbox{{\it #1\/}}}}
\def\action#1{{\hbox{{\it #1\/}}}}

\def\etal{{\em et al.}\xspace}
\def\rest1.0{{\tt restaurant-1.0}\xspace}

\newcommand{\no}{\mbox{not }}

\def\lth{{\sc lth}\xspace}
\def\core{{\sc coreNLP}\xspace}

\definecolor{codegray}{rgb}{0.5,0.5,0.5}

  \title[ASP Theories of Intentions and Restaurant Stories]
	      {An Application of ASP Theories of Intentions to Understanding Restaurant Scenarios:\\
				Insights and Narrative Corpus}

  \author[Q. Zhang \and C. Benton \and D. Inclezan]
         {QINGLIN ZHANG \and CHRIS BENTON \and DANIELA INCLEZAN\\
         Miami University, College of Engineering \& Computing, Oxford OH 45056, USA\\
         \email{zhangq7,bentoncl,inclezd@miamioh.edu}}

\jdate{March 2003}
\pubyear{2003}
\pagerange{\pageref{firstpage}--\pageref{lastpage}}
\doi{S1471068401001193}

\newtheorem{example}{Example}

\begin{document}

\label{firstpage}

\maketitle

\begin{abstract}
This paper presents a practical application of Answer Set Programming 
to the understanding of narratives about restaurants. 
While this task was investigated in depth by Erik Mueller,
{\em exceptional} scenarios remained a serious challenge for his script-based 
story comprehension system.
We present a methodology that remedies this issue by modeling
characters in a restaurant episode as {\em intentional agents}.
We focus especially on the refinement of certain components of 
this methodology in order to increase coverage and performance.
We present a 
restaurant story corpus that we created 
to design and evaluate our methodology.
Under consideration in Theory and Practice of Logic Programming (TPLP).
\end{abstract}



\section{Introduction}
\label{sec:introduction}
In this paper, we present an application of Answer Set Programming (ASP)
to the understanding of restaurant narratives.
Dining at a restaurant is a stereotypical human activity, i.e.,
{\em a sequence of actions normally performed in a certain order by one or more actors, 
according to cultural conventions.}
Automating a deep understanding of stories about stereotypical human activities 
is a more difficult task than that of understanding other types of narratives
because a larger number of events that are part of the activity are not explicitly mentioned
in the text, with the assumption that readers will be able to fill in the gaps based on 
their shared cultural knowledge. Consider the following example:

\begin{example}[Normal scenario from \cite{m07}]
\label{ex:normal}
{\em Nicole went to a vegetarian restaurant. She
ordered lentil soup. The waitress set the soup
in the middle of the table. Nicole enjoyed the
soup. She left the restaurant.}
\end{example}

\noindent
The story in Example \ref{ex:normal}
does not mention that the waitress went to the kitchen to get the soup nor that 
Nicole paid for her meal, as these actions are implicitly assumed. 

We chose to focus specifically on restaurants with table service because
stories about this domain involve more actors, performing more actions, and
interacting in more complex ways than in other types of restaurants
(e.g., fast food restaurants). As a consequence, knowledge representation and reasoning
techniques become relevant to understanding stories about the chosen stereotypical
activity, while an automated learning of the sequence of events that
form this activity would be very difficult, as indicated in Section \ref{sec:relatedWork}.
Our work is applicable to stories about other stereotypical activities.

\smallskip
Schank and Abelson \citeyear{sa77} proposed modeling stereotypical human activities as {\em scripts}, i.e.,
``standardized sequences of events'' \cite{bf81},
and Mueller conducted substantial research in this direction (e.g., \cite{m04}),
including work on the restaurant domain specifically \cite{m07}. However,
his system was not able to understand stories describing exceptional scenarios like the ones in 
Examples \ref{ex:serendipity} and \ref{ex:diagnosis} below because of the rigid structure of scripts --
actions in the script are assumed to always occur in the exact order specified in the script.

\begin{example}[Serendipity]
\label{ex:serendipity}
{\em Nicole went to a vegetarian restaurant. She ordered lentil soup.
When the waitress brought her the soup, she told Nicole that it was on the house.
Nicole enjoyed the soup and then left.} 
(The reader should understand that Nicole did not pay for the soup.)
\end{example}

\vspace{-0.4cm}
\begin{example}[Diagnosis]
\label{ex:diagnosis}
{\em Nicole went to a vegetarian restaurant. She
ordered lentil soup. The waitress brought her a miso soup instead. }
(The reader is supposed to produce some explanations for what may have gone wrong:
either the waitress or the cook misunderstood the order.)
\end{example}

We have argued \cite{zi17,izbi18} that modeling actors in a restaurant scenario as goal-driven 
intentional agents is needed in order to be able to process exceptional restaurant scenarios
in addition to normal ones. We have proposed to use theories of intentions written in ASP \cite{bg05i} or easily
translatable into ASP \cite{thesisblount13,bgb15} to model the characters 
in a restaurant episode as intentional agents, and 
concluded that our methodology has a wider coverage than script-based approaches.

{\em In this paper, we investigate remaining research questions related to the proposed methodology
and present a 
corpus of restaurant stories that we built in order to evaluate our
methodology, and which we make publicly available.}

Our first research question studies the impact in terms of coverage and performance 
of modeling all characters in a restaurant scenario as goal-driven intentional agents 
as defined by Blount \etal \citeyear{bgb15}
versus viewing only the main character, the customer, as such, and modeling
other characters using a simpler theory of intentions by Baral and Gelfond \citeyear{bg05i}.
The second research question investigates the optimal structure for the representation of
the stereotypical activity of dining at a restaurant from the point of view of each character.
We envision the restaurant corpus that we constructed, \texttt{restaurant-1.0}, as a resource
to be used in future research on stereotypical activities, but also 
a useful benchmark for the NLP, natural language understanding, KRR, and ASP communities.

\smallskip
The contributions of this work are as follows:
\begin{itemize}[leftmargin=*,noitemsep,topsep=0pt]
\item We demonstrate that, by using ASP theories of intentions, 
our proposed methodology can reason about exceptional scenarios that can not be
processed using traditional, script-based approaches.
Thus, we introduce and highlight an important application area for 
the ASP body of work on theories of intentions.
\item We indicate that modeling only the customer role as a goal-driven intentional agent 
presents advantages in terms of performance, while 
having only a moderate negative impact on coverage.
\item We provide guidelines for structuring the representation of stereotypical human activities
based on lessons learned from serendipitous scenarios like the one in 
Example~\ref{ex:serendipity},
which require a hierarchical structure where activities have sub-activities with sub-goals,
and scenarios involving diagnosis like the one in Example \ref{ex:diagnosis},
which require paying attention to the parameters of the activity.
\item We introduce and make available a corpus of restaurant stories 
accompanied by their ASP logic forms.
\end{itemize}

In what follows, we start by discussing related work and then describe
the proposed methodology for reasoning about restaurant stories.
Next, we explore the two research questions connected to our methodology.
We then briefly present the application of our refined methodology on a few illustrative stories. 
We present the restaurant story corpus that we created and
end with conclusions and future work.


\section{Related Work}
\label{sec:relatedWork}

\noindent
{\bf Restaurant Narratives.} 
Erik Mueller's work is based on the hypothesis that readers of a text 
understand it by constructing a mental model \cite{c43} 
of the narrative \cite{jl83,vdk83}.
Mueller's system \citeyear{m07} showed a deep understanding of restaurant
narratives by answering questions about time and space aspects that were not necessarily mentioned
explicitly in the text.
His system relied on two important pieces of background knowledge:
a commonsense knowledge base about actions occurring in a restaurant, 
their effects and preconditions, encoded in Event Calculus \cite{s97};
and 
a script describing a sequence of actions performed 
in a {\em normal} unfolding of a restaurant episode. The script was much more detailed than those
used in other systems, for instance Ng and Mooney's plan recognition software ACCEL \citeyear{nm92},
and thus was able to demonstrate a more in-depth understanding.

The system processed English text using information extraction techniques 
in order to fill out slot values in a template. 
Table \ref{table1} shows the template constructed for the scenario in Example \ref{ex:normal}.
Note that the slot SCRIPT: LAST EVENT is filled with the value {\sc Leave},
which corresponds to the customer's last action in the restaurant script. 
\begin{table}[!hbt]
\caption{Slot values for the template constructed for Example \ref{ex:normal}}
\label{table1}
\begin{minipage}{\textwidth}
\begin{tabular}{l l}
\hline\hline
{\bf Slot} & {\bf Slot Value}\\
\hline
{\bf SCRIPT: TYPE} & {\sc Restaurant}\\
{\bf SCRIPT: LAST EVENT} & {\sc Leave}\\
{\bf RESTAURANT} & ``the vegetarian restaurant''\\
{\bf WAITER} & ``waitress''\\
{\bf CUSTOMER} & ``Nicole''\\
{\bf FOOD} & ``lentil soup'' : ``Nicole''\\
\hline\hline
\end{tabular}
\end{minipage}
\end{table}
Next, the template was translated into a reasoning problem 
that contained: facts about the entities identified in the template;
facts about the consecutive occurrence of all actions in the script
up to the one corresponding to the value of the SCRIPT: LAST EVENT slot;
and
default information about the layout of the restaurant and the locations
of different objects and characters.
Then, the reasoning problem was expanded with the information in the commonsense knowledge base
to compute models of the input restaurant scenario.
Finally, questions about time and space aspects were automatically generated  
and answers were obtained from the model resulting in the previous step.
Mueller's system was tested on 124 excerpts of texts retrieved from the
web or Project Gutenberg collection,
and answered correctly 70\% of the test questions. 

In terms of limitations of the system, the author acknowledged
the lack of flexibility of scripts, which resulted in scenarios with exceptional cases (or
variations, such as an additional wine tasting step) not being processed correctly. 
For instance, for the scenario in Example \ref{ex:serendipity},
Mueller's system would detect {\sc Leave} as the SCRIPT: LAST EVENT and thus construct the same template 
as the one for Example \ref{ex:normal}. As a result, the reasoning problem built from the template
would include the fact that Nicole paid for the soup, since this action
precedes the customer's action of leaving in the script, 
when in fact a human reader would infer that she did not pay because the soup was on the house.
In the script-based approach, such a serendipitous scenario can only be solved by introducing 
a new script that does not contain the pay action.
This means that the knowledge engineer would have to predict all possible exceptional scenarios and 
create a new script for each of them in advance.

\medskip
\noindent
{\bf Narrative Corpora.}
Mueller's two restaurant story corpora \citeyear{m07}, one based on Internet stories 
and the other on Project Gutenberg texts, are proprietary and thus unavailable.
Reconstructing these corpora is a laborious task.
General story corpora exist, but they do not apply to the subject of this research.
For instance, the InScript narrative corpus \cite{maop17} covers other stereotypical human activities 
(e.g. grocery shopping, taking the bus),
but not the topic of dining in a restaurant with table service.
The OMCS (Open Mind Common Sense) \cite{slmlpz02} and 
OMICS (Open Mind Indoor Common Sense) \cite{guptak04} corpora
cited in earlier papers, though publicly available in the past, are no longer readily available and 
do not contain 
restaurant stories. 
The SMILE corpus \cite{rkp10} contains stories about eating at a fast food restaurant,
but not at an elegant restaurant.
Gordon {\em et al.} \citeyear{gcs07} processed and annotated an existing corpus of
stories extracted from Internet web blogs.
However, due to a complex agreement system for using the original web blog corpus, 
the data is not readily available to the public.

\medskip
\noindent
{\bf Automated Learning of Activities.}
In recent years, there has been an increased interest in automatically learning the 
sequence of events that forms a stereotypical activity \cite{cj08,msg08,rkp10}.
However, these approaches are only able to produce flat sequences of actions that
are not associated with goals. Smith and Arnold \citeyear{sa09} are able to produce
hierarchical plans, but these are not associated with goals either.
As stated in the introduction, a hierarchical structure 
with sub-sequences and associated (sub-)goals is required for a system to be able to 
reason about exceptional scenarios like the ones in Examples \ref{ex:serendipity} and \ref{ex:diagnosis}.
Additionally, the targeted stereotypical activities in this 
unsupervised learning body of work do not include dining at a restaurant with table service
and generally have only one actor (e.g., make coffee).

\medskip
\noindent
{\bf Activity Recognition.} 
The task of automating the understanding of restaurant narratives is somewhat connected 
to activity recognition, in that it requires observing agents and their environment
in order to complete the picture about the agents' actions and activities.
However, unlike activity recognition, understanding restaurant narratives 
does not require identifying an agent's goal, which is always the same in our
case (e.g., a customer entering a restaurant always seeks to become satiated).
Gabaldon \citeyear{Gabaldon09} performed activity recognition using 
Baral and Gelfond's theory of intentions \citeyear{bg05i} 
that did not consider goal-driven agents.

\medskip
\noindent
{\bf Preliminary Work.}
In an earlier version of this paper, Zhang and Inclezan \citeyear{zi17} 
presented an initial solution to the problem of reasoning about restaurant stories 
by using both of the existing theories of intentions.
Inclezan \etal \citeyear{izbi17,izbi18}
introduced the alternative approach of using the newer theory of intentions for modeling all
characters in a restaurant story. Neither of these papers compare the two approaches nor include
the work on the restaurant story corpus.


\section{Reasoning about Restaurant Stories in ASP}
\label{sec:reasoning}

As mentioned in the introduction, the script-based approach is not suitable
for reasoning about exceptional scenarios because of the rigidity of scripts.
In previous work \cite{zi17,izbi17,izbi18}, we proposed a new approach, 
capable of handling normal and exceptional scenarios,
based on the idea of viewing the main character in a restaurant scenario (and possibly others) 
as a goal-driven intentional agent.
We used theories of intentions to reason about the actions of such agents,
coupled with a background knowledge base about actions and properties (fluents) relevant to the restaurant domain,
as well as an encoding of the stereotypical activity itself, from the point of view of each character.
In this section, we briefly introduce the two existing theories of intentions, written in ASP or languages closely related to ASP.
We then outline our methodology and stress specifically the research questions that resulted 
from our preliminary work.

\subsection{Theory of Intended Actions by Baral and Gelfond}

Baral and Gelfond \citeyear{bg05i} captured properties of intended actions 
in an ASP theory we denote by \oldTI that had two main
tenets: {\em ``Normally intended actions are executed the moment
such execution becomes possible''} (non-procrastination) and
{\em ``Unfulfilled intentions persist''} (persistence).
Sequences of actions were modeled using predicates:
$sequence(\var{s})$ ($\var{s}$ is a sequence); 
$length(\var{n}, \var{s})$ ($\var{n}$ is the length of sequence $\var{s}$); and
$component(\var{s}, \var{k}, \var{x})$ (the $\var{k}^{th}$ element of sequence
 $\var{s}$ is $\var{x}$, where $\var{x}$ can be either an action or another sequence).
\noindent
An agent's intentions at different time points was captured by 
$intend(\var{x}, \var{i})$ (action/ sequence $\var{x}$ is intended at time step $\var{i}$).
The theory was successfully used in activity recognition \cite{Gabaldon09} 
and question answering about biological processes \cite{ig11},
but was not sufficient for modeling goal-driven agents.
We use the term {\em simple intentional agent} to refer to an agent
that can be modeled by \oldTI.

\subsection{Theory of Goal-Driven Intentional Agents by Blount \etal}

Blount and collaborators \cite{thesisblount13,bgb15} 
improved on the previous theory of intentions 
by considering goal-driven agents inspired by the 
Belief-Desire-Intention (BDI) model \cite{b87}. For this purpose, 
each sequence of actions of an agent was associated 
with a goal that it was meant to achieve -- the combination of the two
was called an {\em activity}. Activities could have nested sub-activities, and
were encoded using the predicates:
$activity(\var{m})$ ($\var{m}$ is an activity); 
$goal(\var{m}, \var{g})$ (the goal of activity $\var{m}$ is $\var{g}$); 
$length(\var{n}, \var{m})$ (the length of activity $\var{m}$ is $\var{n}$); and
$component(\var{m}, \var{k}, \var{x})$ (the $\var{k}^{th}$ component of activity 
$\var{m}$ is $\var{x}$, where $\var{x}$ is either an action or a sub-activity). 

The authors introduced the concept of a {\em goal-driven intentional agent} --- one that 
has goals that it intends to pursue, ``only attempts
to perform those actions that are intended and does so without delay.''
To represent the intentions and decisions of an intentional agent, 
Blount {\em et al.} introduced 
{\em mental} fluents and actions.
Two important mental fluents 
are $status(\var{m}, \var{k})$ ($\var{m}$ is in progress if
$\var{k} \geq 0$, and not yet started or stopped if $\var{k}=-1$)
and $next\_action(\var{m}, \var{a})$ (the next action to be executed as part of 
activity $\var{m}$ is $\var{a}$). 
Mental actions included $select(\var{g})$ and $abandon(\var{g})$ for goals,
and $start(\var{a})$ and $stop(\var{a})$ for activities.
The new theory of intentions was encoded in action language ${\cal AL}$ \cite{bg00}.
We denote by \newTI its ASP translation.

Additionally, Blount {\em et al.} developed an agent architecture \AIA
(implemented in CR-Prolog \cite{bg03a,b07}, an extension of ASP) 
that adapts the agent loop \cite{bg08} to specify the behavior of a goal-driven intentional agent.
For instance, while fluent $next\_action(\var{m}, \var{a})$ in the theory of intentions 
indicates the action in activity $\var{m}$ that the agent would {\em normally}
 need to execute next,
the agent architecture handles exceptions to this rule. 
The decision not to execute the next action is made 
if the activity's goal was already achieved by 
some other action (Example \ref{ex:serendipity}) or 
was abandoned; or if
the current activity needs to be stopped altogether because it no longer has chances
of achieving its goal (Example \ref{ex:diagnosis}).


\subsection{Proposed Methodology}
\label{sec:methodology}

Our methodology describes how to construct an ASP logic program for each input restaurant narrative
based on the information given in the text, a background commonsense knowledge base,
theories of intentions, and an adapted and extended version of the \AIA architecture.
Answer sets of the resulting program correspond to a cautious reader's possible mental models
of the narrative, which can be used to demonstrate a deep understanding of the story via question answering.
By ``deep understanding'' we mean awareness of the intentions of characters and of the occurrence
of actions that were not explicitly stated in the text but would be assumed by a human reader.

Our goal is to focus on the reasoning component.
We thus ignore the natural language processing part, 
which is a difficult task on its own.
We distinguish between the story time line containing strictly the events mentioned in the text
and the reasoning time line corresponding to the mental model that the reader constructs.
We assume that a wide coverage commonsense knowledge base (\kb) written in ASP 
is available to us and that it contains information about a large number of 
actions, their effects and preconditions, including actions in the stereotypical activity.
How to actually build such a knowledge base is a difficult research question, but
it is orthogonal to our goal. In practice, in order to be able to evaluate our methodology, 
we have built a basic knowledge base with core information 
about restaurants in the spirit of Mueller's work \citeyear{m07}. 

According to our methodology, for each input text $t$ we construct a logic program $\Pi(t)$ consisting of 
an input-dependent part 
and a pre-defined part common to all texts.

\medskip
\noindent
{\bf The input-dependent part  of $\Pi(t)$} (i.e., the logic form obtained 
by translating the English text $t$ into ASP facts) 
consists of facts defining objects mentioned in the text 
as instances of relevant sorts in the \kb and 
observations about the values of fluents and the occurrences of actions at different points on the 
{\em story} time line. To record observations about fluents and actions, we use predicates $st\_obs$ and $st\_hpd$ respectively.
By $st\_obs(\var{f}, \var{v}, \var{ss})$ 
we mean that fluent \var{f} from the \kb has value \var{v} at time step \var{ss}
on the story time line, where \var{v} may be \inst{true} or \inst{false}), and 
$st\_hpd(\var{a}, \var{v}, \var{ss})$ 
indicates that action \var{a} from the \kb was observed to have occurred if \var{v} is \inst{true}, or 
not if \var{v} is \inst{false}, at
time step \var{ss} on the story time line. 
Let us illustrate the logic form obtained for a sample scenario.

\begin{example}[Input-dependent part (i.e., logic form) for story in Example \ref{ex:normal}]
\label{lf_ex}
The text in Example \ref{ex:normal} is translated into a logic form
that includes the following facts:

\smallskip
\begin{tabular}{l l}
$
\begin{array} {l}
customer(\inst{nicole}).\\
restaurant(\inst{veg\_r}).\\
food(\inst{lentil\_soup}).\\
waitress(\inst{waitress}).\\
cook(\inst{cook1}).
\end{array}
$
&
$
\begin{array}{l}
st\_hpd(enter(\inst{nicole}, \inst{veg\_r}), \inst{true}, 0).\\
st\_hpd(order(\inst{nicole}, \inst{lentil\_soup}, \inst{waitress}), \inst{true}, 1).\\
st\_hpd(put(\inst{waitress}, \inst{lentil\_soup}, \inst{t}), \inst{true}, 2).\\
st\_hpd(eat(\inst{nicole}, \inst{lentil\_soup}), \inst{true}, 3).\\
st\_hpd(leave(\inst{nicole}), \inst{true}, 4).
\end{array}
$
\end{tabular}
\noindent
where $enter$, $order$, $put$, $eat$, and $leave$
are actions described in \kb
\end{example}

\medskip
\noindent
{\bf The pre-defined part of $\Pi(t)$} consists of:

\begin{enumerate}[leftmargin=*,noitemsep,topsep=0pt]
\item[1.] The background commonsense {\bf knowledge base \kb}, which contains information about actions and fluents relevant to the restaurant domain,
including axioms about the direct, indirect effects and preconditions of actions. 
These are encoded in ASP using a standard methodology \cite{gk14}
in which predicates $holds(\var{f}, \var{i})$ and
$occurs(\var{a}, \var{i})$ denote the beliefs that fluent $\var{f}$ holds at time step $\var{i}$
and action $\var{a}$ occurs at $\var{i}$ respectively. For example, the two rules
below encode one direct effect and one executability condition for action $put(\var{p}, \var{t}, \var{l})$ --
person \var{p} puts thing \var{t} on location \var{l}:

$
\begin{array}{lll}
\ \ \neg holds(holding(P,T),I+1) & \leftarrow & occurs(put(P,T,L),I),\ holds(holding(P,T),I).\\
%
\ \ impossible(put(P, T, L), I) & \leftarrow & location(L),\ \neg holds(holding(P, T), I).		
\end{array}
$					

%
%

\item[2.] {\bf A theory (or theories) of intentions.} 

\item[3.] A module {\bf encoding the stereotypical activity} from the perspective of each actor.

\item[4.] {\bf A reasoning module}, encoding (i) a mapping of time points on the story time line
into points on the reasoning time line; and (ii) reasoning components 
that reflect a reader's reasoning process and
expected to allow reasoning about serendipitous achievement of goals,
decisions to stop futile activities, and diagnosis.

(i) To encode the mapping of story time steps to reasoning time steps we introduce 
the predicates $story\_step$ and $step$, respectively, as well as the predicate
$map(\var{s}, \var{i})$ to say that story step \var{s} is mapped into reasoning time
step \var{i}:

$
\begin{array}{lll}
\ \ 1 \{map(S, I) : step(I) \} 1 & \leftarrow & story\_step(S).
\end{array}
$

$
\begin{array}{lll}
\ \ \neg map(S, I) & \leftarrow &   map(S_1, I_1),\ S < S_1,\ I \geq I_1,\ story\_step(S),\ step(I).
\end{array}
$

%
%
%

Observations about the occurrence of actions and values of fluents, recorded
from the text using predicates $st\_hpd$ and $st\_obs$, are translated
into observations on the reasoning time line, for which we use the predicates
$hpd$ and $obs$, via rules of the type:

$
\begin{array}{l}
\ \ hpd(A, V, I) \ \leftarrow \ st\_hpd(A, V, S),\ map(S, I).
\end{array}
$

\noindent
Finally, ``gaps'' on the reasoning time line are prevented by the rules 

$
\begin{array}{l}
\ \ smtg\_occurs(I) \ \leftarrow \ occurs(A, I).\\
\ \ \ \leftarrow \ last\_assigned(I),\ step(J),\ J < I,\ \mbox{not }smtg\_occurs(J).
\end{array}
$

\noindent
where $last\_assigned(\var{i})$ is true if \var{i} is the last time step
on the reasoning time line that has a correspondent on the story time line.

(ii) Reasoning components are adapted from \AIA and expanded to
reflect the reasoning process of an outside observer (the cautious reader)
instead of that of an agent thinking about its next action.
For instance, \AIA rules indicating how an agent should select a new activity
to satisfy an active goal if the current activity is deemed futile are replaced by a single
rule indicating that the mental action of replanning has occurred:

$
\begin{array}{lll}
\ \ occurs(replan(Ag, G),I) 
     & \leftarrow & categ\_4\_hist(G,I), \ \mbox{not } futile(Ag, G, I),\\
     & & \mbox{not }impossible(replan(Ag, G), I).
\end{array}
$

The cautious reader is not expected to guess
what new activity the agent decided to start, unless this is explicitly specified in the text.

\item[5.] {\bf Default information} about the values of fluents in the initial situation
(e.g., the restaurant is normally open, dishes listed on the menu are normally available, etc.)
\end{enumerate}

\smallskip
The proposed methodology was tested with good results in previous work: in one instance, only the
customer role was modeled as a goal-driven agent using \newTI while other actors were modeled as
simple intentional agents using \oldTI \cite{zi17}; 
in the other case, all characters were modeled as 
goal-driven agents using the \newTI \cite{izbi18}.
However, a couple of important research questions still remain about components 2 and 3 of the pre-defined 
part:
\begin{itemize}
\item[$RQ_1$]
If we were interested in answering questions about the goals and intentions of the customer only,
what are the trade-offs in terms of coverage and performance between
viewing only the customer as a goal-driven intentional agent 
(i.e., using the \newTI for the customer only and the \oldTI for all other characters --
case denoted by \newoldTI) 
versus
viewing all characters as goal-driven agents 
(i.e., using the \newTI for all of the characters -- case denoted by \newTIonly)?
\item[$RQ_2$]
How should we structure the representation of the stereotypical activity, from the point of view
of each actor, in order to maximize coverage and performance?
\end{itemize}

\noindent
We define {\em coverage} as the number of different {\em types} of scenarios
that can be processed correctly. Scenario types include: stories with only one customer versus multiple
customers, plus the different scenario types listed in Baral \etal's \citeyear{bgb15}
work on theory of intentions
(e.g., normal, serendipitous, diagnosis scenarios).

In the next section we present our insights into these two research questions.


\section{Research Questions: Insights}


\subsection{Insight \#1: Two Theories of Intentions}
\label{sec:twoTIs}

We start by focusing on research question $RQ_1$ and analyze the two cases listed above:
\newoldTI and \newTIonly.
The case of viewing all characters as simple intentional agents and thus using only \oldTI 
is not an option. This approach is too limiting and does
not allow reasoning correctly about scenarios like the one in Example \ref{ex:serendipity}
for reasons similar to those related to the script-based approach. 

\subsubsection{Coverage} 
The advantage of viewing a character as a goal-driven intentional agent 
(and using the \newTI instead of the \oldTI to model the character's activity) 
is that it allows reasoning about the serendipitous achievement of the character's goals and sub-goals. 
This means that using the \oldTI instead of \newTI for secondary characters (e.g., the waiter)
leads to scenarios like the one in Example \ref{ex:ser_waiter} 
not being processed correctly: 

\begin{example}[Serendipity for Waiter]
\label{ex:ser_waiter}
{\em Nicole went to a vegetarian restaurant. She ordered a lentil soup. Nicole was in a hurry,
so as soon as the waitress laid the dish on the table, Nicole paid for it in cash
and said that she didn't need the bill.}
(The reader is expected to understand that the waitress did not bring the bill to Nicole.)
\end{example}

\noindent
For the story in Example \ref{ex:ser_waiter}, an answer set would be produced by the \newoldTI approach, but
it would inaccurately state that the waitress did bring the bill to Nicole. 
This is a drawback for the \newoldTI case.

\smallskip
On the other hand, the \newoldTI case allows reasoning about scenarios involving multiple
customers, each ordering a different dish as in Example \ref{ex:multiple_cust}:

\begin{example}[Multiple Customers]
\label{ex:multiple_cust}
{\em Nicole and Sam went to a vegetarian restaurant. She ordered a lentil soup. 
He ordered a miso soup. They both enjoyed their soups.}
\end{example}

In our formalization of the domain, the waiter either maintains an \oldTI sequence of actions
for each customer (case \newoldTI) or, alternatively, it maintains one \newTI activity 
per customer, with the associated goal of serving and billing the customer (case \newTIonly).
However, a waiter cannot maintain multiple \newTI activities at a time, corresponding
to multiple customers, because of a current limitation in Blount \etal's theory of intentions 
indicating that an agent can only have one top-level active goal at a time.
As a result, applying the \newTIonly approach to such scenarios would result in no answer sets.
Substantial work on goal selection and prioritization 
is needed in order to lift this restriction.
With the \newoldTI solution, the secondary role of waiter is modeled 
as a simple intentional agent
who does not maintain goals, but rather follows a sequence of intended actions. 
As a result, the waiter may entertain multiple sequences of actions at a time.

In terms of the scenarios that we found while working on the restaurant narrative corpus,
a higher number of examples involved multiple customers in the style of
Example~\ref{ex:multiple_cust} (27.5\%) than
those requiring to view secondary characters as goal-driven intentional agents 
as in Example~\ref{ex:ser_waiter} (15\%).


\subsubsection{Performance}
The \newTI, which is required to reason about goal-driven intentional agents, is much more complex than the \oldTI.
A comparison in terms of different measures can be seen in Table \ref{tab:comparison}.
This has a substantial impact on the performance of a system implemented 
according to our methodology, especially on input stories that involve diagnosis, 
which is a combinatorial search for an explanation. 

\begin{table}[!htp]
\caption{Comparison between \oldTI and \newTI}
\label{tab:comparison}
\begin{minipage}{\textwidth}
\begin{tabular}{l c c}
\hline\hline
 {\bf Metric} & {\bf \oldTI} & {\bf \newTI}\\
\hline
Number of rules & 6 & 111 \\
Number of lines of code & 45 & 453\\ 
Number of fluents & 1 & 12 \\
Number of actions & 0 & 2 \\
\hline
Minimum number of steps on the reasoning time line & $n$ & $n+2$\\
for a flat sequence/ activity of length $n$  & &\\
\hline\hline
\end{tabular}
\end{minipage}
\end{table}

Consider for instance the last metric in Table \ref{tab:comparison}. 
If activities are represented using a hierarchical structure with sub-activities that have associated sub-goals
(which is desired, as we will show in the next subsection),
then each sub-activity adds two additional time steps on the reasoning time line:
one for the mental action of starting the sub-activity and another one for stopping the sub-activity.
This happens even when the sub-activity's goal is serendipitously satisfied by some other agent's actions
and none of the physical actions in the sub-activity are performed by the agent (see the output for 
Example \ref{ex:serendipity} in Section \ref{sec:eval}).
Moreover, no physical actions of the same agent can occur while 
a mental action is happening, and some restrictions about physical actions of other agents also exist.
A larger number of steps on the reasoning time line has an impact on diagnosis problems 
especially, as shown in Table \ref{tab:comparison_sc}. 
The reported times are the averages of ten runs 
on a machine with an 
Intel(R) Core(TM) i5-4300U CPU \@1.9GHz and 4GB RAM
using the {\sc clingo}4.5.4 solver\footnote{https://sourceforge.net/projects/potassco/files/clingo/}.

\begin{table}[!htp]
\caption{Performance comparison. (By {\em Max Step} 
we denote the maximum time step on the reasoning time line when 
an action is considered to have occurred.)}
\label{tab:comparison_sc}
\begin{minipage}{\textwidth}
\begin{tabular}{l | c r | c r | c }
\hline\hline
\multirow{2}{*}{{\bf Scenario}} & \multicolumn{2}{c |}{\newoldTI} 
                                & \multicolumn{2}{ c}{\newTIonly} 
                                & {} \\ 

\multirow{2}{*}{} & Avg. Time & {\em Max Step\ } & Avg. Time  & {\em Max Step\ } & Time increase\\
\hline
Normal & 1.07s  & {\em 29\ } & 1.61s & {\em 33\ } & 50.74\%\\
Serendipity & 1.91s & {\em 23\ } & 2.52s & {\em 27\ } & 32.25\%\\
Futile Activity  & 1.03s & {\em 8\ } & 1.34s & {\em 9\ } & 29.89\%\\
Diagnosis 1 (wrong dish) & 2.34s & {\em 16\ } & 3.62s & {\em 20\ } & 54.80\%\\
Diagnosis 2 (wrong bill) & 1.28s & {\em 23\ } & 1.95s & {\em 28\ } & 51.98\%\\
\hline\hline
\end{tabular}
\end{minipage}
\end{table}

\noindent
{\bf Answer for $RQ_1$.} Based on this analysis, we conclude that 
\newoldTI has an improved performance over \newTIonly and has 
the potential for a wider coverage as it 
can handle a larger number of what seem to be recurrent scenarios.


\subsection{Insight \#2: Hierarchical Activity Representation}
\label{sec:activity}
 
To answer research question $RQ_2$ about the most suitable structure for the representation
of each actor's actions as part of the stereotypical activity, 
we started from the flat and fixed scripts described in the 
existing body of literature on narratives about restaurants with table service (e.g., \cite{sa77,m07}),
which we then refined.
We considered two main factors that impact decisions about activity structure: 
(1) in order to be able to reason about serendipitous scenarios, 
activities must have a hierarchical structure with sub-activities having their own sub-goals; and
(2) in order to be able to process scenarios that require diagnosis (e.g., wrong dish / bill)
additional parameters may be needed (e.g., one parameter indicating the ordered dish and
another one for the actual, possibly wrong, dish brought by the waiter).
In what follows, we describe our conclusions related to such decisions,
and especially their impact on coverage and performance.
We adopt the conclusion from Section \ref{sec:twoTIs} and assume that the
customer's actions are represented as an activity of \newTI,
while the waiter and cook intend to execute sequences of actions of \oldTI.

\subsubsection{Activity Structure and Serendipitous Scenarios}
In our methodology, reasoning about serendipitous scenarios is possible 
whenever the customer's actions whose purpose 
is satisfied by someone else's actions
are grouped into a sub-activity associated with a goal.
For instance, Example \ref{ex:serendipity} can be processed correctly
if and only if the customer's activity contains a sub-activity consisting of the
payment-related actions (request bill and pay bill) and associated with a goal
that can be satisfied by another character's actions.
When this is the case, the rules in \newTI indicate that the
customer performs the mental action of starting the payment sub-activity, realizes that the goal is already
met, and then performs the mental action of stopping the sub-activity, without performing any of the
physical actions in it. 
To increase the coverage of different serendipitous scenarios, we must make sure that
we create a hierarchical structure for the customer's activity in which all goals
that may be satisfied by other actors' actions are represented and associated with 
a corresponding sub-goal, thus rendering sub-activities optional. 
This is the criterion that we employ to divide the customer's activity into sub-activities, 
of course in addition to grouping together the actions that are intuitively part of the same sub-plan 
(e.g., picking up the menu and putting it back on the table are part of the 
sub-plan of deciding what to order).

One possibility that would guarantee maximum coverage is to package each action into a sub-activity
with a sub-goal as in Activity Theory \cite{bb03}, and then build other sub-activities from there. 
However, this would be detrimental in terms of performance. 
Each new sub-activity that is introduced
adds two mental actions (\action{start} and \action{stop})
that need to be executed by the actor, 
which translates into two additional time points on the reasoning time line 
given that no other actions, physical or mental, can be executed by the actor at the same time.
As a result, this approach would roughly triple the length of the reasoning time line, 
as compared to a flat activity, which will
negatively impact the code that maps story time line steps onto the reasoning time line,
as well as scenarios with diagnosis, as shown in the previous section.
As an example, consider the structures shown in Table \ref{tab:structure},
where $S_1$ only introduces one sub-activity compared to $S_{flat}$.
The average time over ten runs for processing a normal scenario using the \newoldTI approach 
is 0.57s for $S_{flat}$, 0.70s for $S_1$ (22\% increase), and 1.07s for $S_2$ (87\% increase).

There is an obvious trade-off between coverage and performance that impacts the activity structure 
we choose. We decided where to draw the line based on the exceptional scenarios in our restaurant corpus
that were not hand-crafted. We identified as optimal the activity structure $S_2$ shown in 
Table \ref{tab:structure}, which includes sub-activities for 
the customer getting ready to eat ($c\_subact\_r(C, R, W, F)$),
customer deciding what to order ($c\_subact\_o(C, F, W)$)
and customer paying the bill ($c\_subact\_p(C, W)$).
Note that $c\_subact\_r(C, R, W, F)$ is optional in a scenario where the wrong dish is brought by the waiter
and the customer decides to eat it -- at this point the customer drops his initial
intention of eating the original dish and starts a new activity of eating the wrong dish,
but all the actions up to eating (e.g., sit) become irrelevant and should be made optional.

\begin{table}[!htp]
\caption{Possible activity structures for the customer role.}
\label{tab:structure}
\begin{minipage}{\textwidth}
\begin{tabular}{l | l}
\hline\hline
\boldmath$S_{flat}$
&  
\begin{oldtabular}{l}
\begin{oldtabular}{ l l }
{\bf Name:} & \boldmath$c\_act(C, R, W, F)$ \\
{\bf Plan:} & $[\ enter(C, R),\ lead\_to(W, C, \inst{t}),\ sit(C),\ 
                 pick\_up(C, \inst{m}, \inst{t}),\ put(C, \inst{m}, \inst{t}),$\\ 
	          & $\ \ order(C, F, W),\ eat(C, F),\ request(C, \inst{b}, W),\ pay(C, \inst{b}), \ stand\_up(C),$\\
						& $\ \ move(C, \inst{t}, \inst{entrance}),\ leave(C)\ ]$\\
{\bf Goal:}     & $satiated\_and\_out(C)$   
\end{oldtabular}
\end{oldtabular}
\\
\hline
\boldmath$S_{1}$
&  
\begin{oldtabular}{l}
\begin{oldtabular}{ l l }
{\bf Name:} & \boldmath$c\_act(C, R, W, F)$ \\
{\bf Plan:} & $[\ enter(C, R),\ lead\_to(W, C, \inst{t}),\ sit(C),\ 
                 pick\_up(C, \inst{m}, \inst{t}),\ put(C, \inst{m}, \inst{t}),$\\ 
	          & $\ \ order(C, F, W),\ eat(C, F),\ \mathbf{c\_subact\_p(C, W)}, \ stand\_up(C),$\\
						& $\ \ move(C, \inst{t}, \inst{entrance}),\ leave(C)\ ]$\\
{\bf Goal:}     & $satiated\_and\_out(C)$   
\end{oldtabular}
\\
\begin{oldtabular}{ l l}
\ \ \ \ &
\begin{oldtabular}{l l r}
\hline
Name:  & $\mathbf{c\_subact\_p(C, W)}$  & \ \ \ \ \ \ \ \ \ \ \ \ \ \ \ \ \ \ \ \ \ \ \ \ \ \ \ \ \ \ \ \ \ (sub-activity)\\
Plan:  & $[\ request(C, \inst{b}, W),\ pay(C, \inst{b})\ ]$ &\\
Goal:  & $done\_with\_payment(C)$ &
\end{oldtabular}
\end{oldtabular}
\end{oldtabular}
\\
\hline
\boldmath$S_2$
& 
\begin{oldtabular}{l}
\begin{oldtabular}{ l l }
{\bf Name:} & \boldmath$c\_act(C, R, W, F)$ \\
{\bf Plan:} & $[\ \mathbf{c\_subact\_r(C, R, W, F)}, eat(C, F),\ \mathbf{c\_subact\_p(C, W)}, \ stand\_up(C),$\\
            & $\ \ move(C, \inst{t}, \inst{entrance}),\ leave(C)\ ]$\\
{\bf Goal:}     & $satiated\_and\_out(C)$   
\end{oldtabular}
\\
\begin{oldtabular}{ l l}
\ \ \ \ &
\begin{oldtabular}{l l}
\hline
Name:      & $\mathbf{c\_subact\_r(C, R, F, W)}$ \ \ \ \ \ \ \ \ \ \ \ \ \ \ \ \ \ \ \ \ \ \ \ \ \ \ \ \ \ \ \ \ \ \ \ \ \ \ \ \ \ \  (sub-activity)\\
Plan:      & $[\ enter(C, R),\ lead\_to(W, C, \inst{t}),\ sit(C),\ \mathbf{c\_subact\_o(C, F, W)}\ ]$\\
Goal:      & $ready\_to\_eat(C)$
\\ 
\ \ \ &
\begin{oldtabular}{l l}
\hline
Name:      & $\mathbf{c\_subact\_o(C, F, W)}$ \ \ \ \ \ \ \ \ \ \ \ \ \ \ \ \ \ \ \ \ \ \ \ \ \ \ \ \ \ (sub-sub-activity)\\
Plan:      & $[\ pick\_up(C, \inst{m}, \inst{t}),\ put(C, \inst{m}, \inst{t}),\ 
                 order(C, F, W)\ ]$ \\
Goal:      & $order\_transmitted(C)$
\end{oldtabular}
\\ 
\hline
Name:  & $\mathbf{c\_subact\_p(C, W)}$ \ \ \ --- as defined in \boldmath$S_2$\ \ \ \ \ \ \ \ \ \ \ \ \ \ \ \ \ \ \ \ \ (sub-activity)
\end{oldtabular}
\end{oldtabular}
\end{oldtabular}
\\
\hline\hline
\end{tabular}
\end{minipage}
\end{table}

\subsubsection{Activity Parameters and Scenarios Requiring Diagnosis}
We wanted our methodology to be able to encompass exceptional scenarios that require diagnosis
(e.g., the waiter brings the wrong dish/bill). We illustrate our analysis and decision process
on the waiter role. Recall from Section \ref{sec:twoTIs} that we decided to model the waiter 
as a simple intentional agent that follows 
a sequence of intended actions according to the tenets captured by \oldTI. 

For each input text, our program generates all possible sequences of actions that the waiter
may have intended and determines the actual one based on the events described in the story.
For instance, for the story in Example \ref{ex:diagnosis} that mentions both a lentil soup and a miso soup,
our program assumes that the waitress may be executing a sequence of actions that includes 
serving either a lentil soup to Nicole or a miso soup.
Given that the story indicates that the waitress brought the miso soup, the program
will determine that the waitress is executing the latter sequence.
We distinguish between one sequence and the other via parameters of the waiter sequence.
The dish that is actually served must be one of them as captured by the 
third parameter of the waiter sequence in (\ref{w1}) below.
 
In order to account for possible mistakes in communication as well as for 
careless waiters that do not check the customer's order
before serving the dish, we must distinguish between the food order that the waiter understands
and the food he serves: $F_1$ and $F_2$ in the improved version of the waiter sequence shown in (\ref{w2}).
Finally, by analyzing the stories in our restaurant corpus, we determined that another recurrent 
mistake is the wrong bill being brought by the waiter.
Thus, we concluded that the waiter sequence must contain an additional parameter representing the bill 
$B$ that the waiter brings as in (\ref{w3}) and which may not be the customer's actual bill.
\begin{gather}
   \text{{$w\_seq(W, C, F)$}}\label{w1} \\
   \text{{$w\_seq(W, C, F_1, F_2)$}}\label{w2}\\
	 \text{{$w\_seq(W, C, F_1, F_2, B)$}}\label{w3}
\end{gather}  

The waiter's sequence $w\_seq(W, C, F_1, F_2, B)$ can be read as
``waiter W understands that customer C ordered food $F_1$, serves food $F_2$ 
and brings bill $B$ to him/her.''
It consists of the following actions:
$\langle \ 
greet(W, C)$, 
$lead\_to(W, C, \inst{t})$, 
$move(W, \inst{t}, \inst{kitchen})$,
$request(W, F_1, \inst{ck})$ ($W$ requests to the cook to prepare food $F_1$), 
$pick\_up(W, F_2, \inst{kitchen})$ ($W$ picks up food $F_2$ from the kitchen),
$move(W, \inst{kitchen}, \inst{t})$,
$put(W, F_2, \inst{t})$,
$move(W, \inst{t}$, $\inst{counter})$,
$pick\_up(W, B, \inst{counter})$ ($W$ picks up bill $B$ from the counter),
$move(W, \inst{counter}, \inst{t})$,
$put(W, B, \inst{t}) \rangle
$. 

Similarly, we define the cook's intended sequence of actions as 
$ck\_seq(Ck, F, W)$, read as ``cook $Ck$ understands that he has to prepare food $F$ for waiter $W$'',
where $F$ is not necessarily the food ordered by the customer.
A cook's sequence $ck\_seq(Ck, F, W)$ consists of a single action:
$\langle prepare(Ck, F, W) \rangle$.

Note that for the scenario in Example \ref{ex:diagnosis}, the program will determine that the customer
is executing the activity shown on the first row of Table \ref{tab:expl},
while the waiter and cook are executing the sequences of actions corresponding to one of the 
possible cases shown in the same table. Non-matching food items in the customer, waiter, and cook's
actions are possible, due to axioms in the \kb that specify that the actions of ordering a food, requesting
a food to be prepared by the cook, preparing the food, and picking up a food from the kitchen 
have non-deterministic effects if a non-agent action (i.e., exogenous action) that we call \action{interference} 
occurs simultaneously. 
For instance, the first axiom below
says that an \action{interference} occurring at the same time as
a customer's action of ordering some food $F$ causes the waiter 
to understand that the customer is asking for a different food than $F$:

$
\begin{array}{l}
1 \{holds(informed(W,F1,C),I+1) :  other\_food(F1, F)\} 1 \ \leftarrow \\  
\ \ \ \ \ \ \ \ \ \ \ \ \ \ \ \ occurs(order(C,F,W),I), occurs(interference, I).
\end{array}
$

$
\begin{array}{lll}
other\_food(F_1, F) & \leftarrow & food(F), food(F_1), F \neq F_1.
\end{array}
$

$
\begin{array}{lll}
holds(informed(W,F,C),I+1) & \leftarrow & occurs(order(C,F,W),I),\\
& &  \neg occurs(interference,I).
\end{array}
$

\begin{table}[!htp]
\caption{Possible explanations for Example \ref{ex:diagnosis}.}
\label{tab:expl}
\begin{minipage}{\textwidth}
\begin{tabular}{l l}
\hline\hline
Customer: & $c\_act(\inst{nicole}, \inst{veg\_r}, \inst{waitress}, \inst{lentil\_soup})$\\
\hline\hline
Case 1: & $w\_seq(\inst{waitress}, \inst{nicole}, \inst{lentil\_soup}, \inst{miso\_soup}, \inst{b})$\\
        & $ck\_seq(\inst{cook1}, \inst{miso\_soup}, \inst{miso\_soup})$\\
			  & Explanation: The cook misunderstood the food request made by the waiter.\\
\hline
Case 2: & $w\_seq(\inst{waitress}, \inst{nicole}, \inst{lentil\_soup}, \inst{miso\_soup}, \inst{b})$\\
        & $ck\_seq(\inst{cook1}, \inst{lentil\_soup}, \inst{miso\_soup})$\\
			  & Explanation: The cook understood the order correctly but prepared the wrong food.\\
\hline
Case 3: & $w\_seq(\inst{waitress}, \inst{nicole}, \inst{miso\_soup}, \inst{miso\_soup}, \inst{b})$\\
        & $ck\_seq(\inst{cook1}, \inst{miso\_soup}, \inst{miso\_soup})$\\
			  & Explanation: The waitress misunderstood the customer's order.\\
\hline				
Case 4: & $w\_seq(\inst{waitress}, \inst{nicole}, \inst{lentil\_soup}, \inst{miso\_soup}, \inst{b})$\\
        & $ck\_seq(\inst{cook1}, \inst{lentil\_soup}, \inst{lentil\_soup})$\\
			  & Explanation: The waitress picked up the wrong order from the kitchen.\\
\hline\hline
\end{tabular}
\end{minipage}
\end{table}

\medskip
\noindent
{\bf Answer for $RQ_2$.} Based on our analysis, we decided to structure 
the customer's activity as a hierarchical one by introducing 
sub-activities and sub-goals that would allow us to reason about a large number of serendipitous scenarios
(see structure $S_2$ in Table~\ref{tab:structure}).
We also added parameters to the waiter and cook's sequences of actions,
named $w\_seq(W, C, F_1, F_2, B)$ and $ck\_seq(Ck, F, W)$ respectively,
to be able to
target diagnosis scenarios. We made sure that the \kb contained axioms about the
non-deterministic effects of certain actions when an \action{interference} exogenous action occurs simultaneously.


%


\section{Exemplification of the Refined Methodology}
\label{sec:eval}
We employed our answers to research questions $RQ_1$ and $RQ_2$ to produce a refined methodology
for the understanding of restaurant scenarios. In this section, 
we include the outputs generated by our methodology for three different types of stories 
and indicate the understanding demonstrated for each of them.

\noindent
{\bf Normal Scenario in Example \ref{ex:normal}.} 
We use this case as a baseline when explaining the output of exceptional scenarios.
The answer set of the program $\Pi(\ref{ex:normal})$ obtained according to our 
refined methodology
contains the $occurs(\var{ma}, \var{i})$ and $intend(\var{s}, \var{i})$ atoms 
shown below, where \var{ma} is a mental action, \var{s} is a sequence of actions, and
\var{i} is a time point on the reasoning time line:

$
\begin{array}{l}
occurs(select(nicole, satiated\_and\_out(\inst{nicole})),0) \\
occurs(start(nicole, c\_act(\inst{nicole},\inst{veg\_r},\inst{waitress},\inst{lentil\_soup})),1) \end{array}
$

$
\begin{array}{l}
occurs(start(nicole, c\_subact\_r(\inst{nicole},\inst{veg\_r},\inst{waitress},\inst{lentil\_soup})),2) \\
intend(w\_seq(\inst{waitress},\inst{nicole},\inst{lentil\_soup},\inst{lentil\_soup}), 4)
\end{array}
$

$
\begin{array}{l}
occurs(start(\inst{nicole},c\_subact\_o(\inst{nicole},\inst{lentil\_soup},\inst{waitress})),7)\\
occurs(stop(\inst{nicole},c\_subact\_r(\inst{nicole},\inst{veg\_r},\inst{waitress},\inst{lentil\_soup})),11)
\end{array}
$

$
\begin{array}{l}
intend(ck\_seq(\inst{cook1},\inst{lentil\_soup},\inst{waitress}),13)\\
occurs(start(\inst{nicole},c\_subact\_p(\inst{nicole},\inst{waitress})),18)
\end{array}
$

$
\begin{array}{l}
occurs(stop(\inst{nicole},c\_subact\_p(\inst{nicole},\inst{waitress})),25)\\
occurs(stop(\inst{nicole},c\_act(\inst{nicole},\inst{veg\_r},\inst{waitress},\inst{lentil\_soup})),29)
\end{array}
$



\smallskip\noindent
{\bf Serendipitous Achievement of Goal in Example \ref{ex:serendipity}.}
The logic form for this scenario is identical to the one for Example \ref{ex:normal} 
shown in Example \ref{lf_ex}, except that the two observations about actions
taking place at time points 2--4 are replaced by

$
\begin{array}{l}
st\_hpd(pay(\inst{owner},\inst{b}), \inst{true}, 2).\ \ 
st\_hpd(put(\inst{waitress}, \inst{lentil\_soup}, \inst{t}), \inst{true}, 3).
\end{array}
$

\noindent
where \inst{owner} is a new instance of sort \sort{people}.
The answer set of $\Pi(\ref{ex:serendipity})$ contains similar $occurs$ atoms to 
$\Pi(\ref{ex:normal})$ 
plus one for action $pay(\inst{owner},\inst{b})$
up to time step 18 when the customer's sub-activity related to the bill payment, 
$c\_subact\_p$, starts. From then on, it contains the following $occurs$ predicates
for mental actions:

$
\begin{array}{l}
occurs(start(c\_subact\_p(\inst{nicole},\inst{waitress})),18)\\   
occurs(stop(c\_subact\_p(\inst{nicole},\inst{waitress})),19)\\
occurs(stop(c\_act(\inst{nicole},\inst{veg\_r},\inst{waitress},\inst{lentil\_soup})),23)
\end{array}
$

\noindent
Thus, the reader of this scenario understands that Nicole has stopped $c\_subact\_2$ 
immediately after starting it because she realized that its goal is already fulfilled.
Based on this answer set, questions about facts
not mentioned in the narrative can be answered correctly:
{\em Did Nicole pay for the soup? (No);}
{\em Did Nicole leave the restaurant? (Yes).}

\smallskip
\noindent
{\bf Diagnosis in Example \ref{ex:diagnosis}.}
The logic form contains the observations:

$
\begin{array}{l}
st\_hpd(enter(\inst{nicole}, \inst{veg\_r}), \inst{true}, 0).\\
st\_hpd(order(\inst{nicole}, \inst{lentil\_soup}, \inst{waitress}), \inst{true}, 1).\\
st\_hpd(put(\inst{waitress}, \inst{miso\_soup}, \inst{t}), \inst{true}, 2).\\
\end{array}
$

\noindent
Program $\Pi$(\ref{ex:diagnosis}) has four answer sets, 
containing explanations on what may have gone
wrong. We illustrate here only one of them, which corresponds to Case 3 in 
Table \ref{tab:expl}: the waitress misunderstood the order
to be for miso soup and the cook followed her request.
This answer set contains the same $occurs$ predicates as $\Pi(\ref{ex:normal})$
up to time step 9, and considers that the following physical actions occurred
at time step 10:

$
\begin{array}{l}
occurs(interference,10)\\
occurs(order(\inst{nicole},\inst{lentil\_soup},\inst{waitress}),10)
\end{array}
$

\noindent
while the following sequences of actions were executed by the waitress and cook:

$
\begin{array}{l}
intend(w\_seq(\inst{waitress},\inst{nicole},{\bf miso\_soup},{\bf miso\_soup})),4)\\
intend(c\_seq(\inst{cook1},{\bf miso\_soup},\inst{waitress}),13)
\end{array}
$



\section{Restaurant Narrative Corpus}
\label{sec:corpus}

Mueller's work \citeyear{m07} is the most extensive investigation on restaurant stories.
For the purpose of training and evaluating his system, Mueller created two corpora: 
a web corpus containing 800 texts downloaded from the Internet and likely to involve dining in a restaurant,
and a Gutenberg corpus obtained by downloading thirty American literature texts 
from the Project Gutenberg archive \footnote{\url{http://www.gutenberg.org/}}
Mueller's corpora are proprietary; attempts to recreate them indicated that 
this task requires substantial human effort.
Due to the lack of available corpora, 
we assumed the task of creating a benchmark collection of restaurant stories,
which resulted in a corpus we call \rest1.0.\footnote{The corpus is available at 
\url{https://ceclnx01.cec.miamioh.edu/~inclezd/Restaurant1.0/}}
We believe that making this corpus available to the research community is important 
in facilitating research and allowing for comparisons between systems. 
The collection can be useful to researchers in a variety of fields 
including the NLP community, or those bridging between the NLP and KRR communities. 

When deciding what stories to include in our corpus, we strove to satisfy the following desired properties, 
adopted from previous work on ASP benchmark set selection \cite{hkss13}:
\begin{itemize}[noitemsep,topsep=0pt]
\item[($P_1$)] {\em Broad selection}, i.e., using a variety of sources for the excerpts;
\item[($P_2$)] {\em Fair selection}, i.e., one source should not dominate other sources 
in terms of representation in the corpus;
\item[($P_3$)] {\em Adapted hardness}, i.e., stories and questions should not be too easy nor too hard 
from the KRR point of view, as well as with respect to the NLP task of 
producing logic forms from natural language texts, which we plan to automate in the future; 
\item[($P_4$)] {\em Free of duplicates, reproducible, and publicly available} (duplication of excerpts
was an issue for Mueller's web corpus).
\end{itemize}

To satisfy desired property $P_1$, we selected excerpts form a variety of sources
for inclusion in the \rest1.0 corpus:
Youtube videos about restaurant scenarios intended for English as a Second Language (ESL) learners,
texts available via Google Books, Project Gutenberg texts, stories from Mueller's paper \citeyear{m07},
and hand-crafted scenarios. Table \ref{tab:corpus_distr} shows the distribution of corpus excerpts.
There is a somewhat balanced representation of sources,
if we ignore excerpts retrieved from Mueller's paper and Project Gutenberg, 
which shows at least a moderate satisfaction of desired property $P_2$.

\begin{table}[!htp]
\caption{Distribution of excerpts in the \rest1.0 corpus}
\label{tab:corpus_distr}
\begin{minipage}{\textwidth}
\begin{tabular}{l | r r r r | r }
\hline\hline
{\bf Source}\  & \multicolumn{4}{c |}{\bf Number of excerpts}
                              & {\bf Percentage}   \\ 
{} & {\em Normal} & {\em Exception} & {\em Variation\ } & {\em {\bf Total\ }} 
                  &  {} \\
\hline
Youtube videos & {\em 6} & {\em 6} & {\em 0} & {\bf 12\ } & 30.0\%        \\
Google Books & {\em 4} & {\em 3} & {\em 5} & {\bf 12\ } & 30.0\%       \\
Project Gutenberg & {\em 0} & {\em 2} & {\em 0} & {\bf 2\ } & 5.0\%        \\
Mueller \citeyear{m07} & {\em 1} & {\em 0} & {\em 0} & {\bf 1\ } & 2.5\%        \\
Hand-crafted & {\em 2} & {\em 11} & {\em 0} & {\bf 13\ } & 32.5\%        \\
\hline
{\bf Total} & {\em 13} & {\em 22} & {\em 5} & {\bf 40\ } & 100.0\%        \\
\hline\hline
\end{tabular}
\end{minipage}
\end{table}

To address property $P_3$, we made sure not to include stories that explicitly mentioned 
less than two restaurant-related events (so that the excerpts would not be too hard)
nor stories that omitted less than three such events (not too easy). 
Given that we focused on the KRR task, we wanted to make sure that the
excerpts in the corpus could be handled with moderate to high accuracy by existing
NLP tools like \lth \cite{lth,lthurl} or \core \cite{corenlpurl}, 
while still sounding natural to a native speaker of English. 
To do so, we adapted the original stories by transforming dialogs into narratives,
extracting only the sentences directly related to the restaurant scenario, and 
simplifying some of them, 
while not removing co-references for instance. 
More details are provided below when 
we describe the process for each source of excerpts.
%

\begin{figure}[!htbp]
\centering
\setlength\fboxsep{0pt}
\setlength\fboxrule{0.5pt}
\fbox{\includegraphics[width=1\textwidth]
{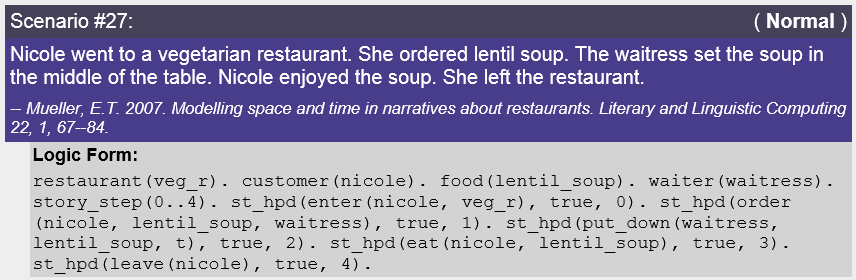}}
\caption{Online display for the \rest1.0 corpus}
\label{fig:corpus}
\end{figure}

To satisfy desired property $P_4$, we made the corpus publicly available.
For each story in the corpus, we recorded the adapted excerpt (what Mueller calls
``condensed narrative''), 
the source, the scenario type (normal/exception/variation), 
and our logic form encoding in ASP according to the description in Section \ref{sec:methodology}. 
We believe that including the manually produced logic forms in the corpus
represents a useful resource for NLP research.
Each story was assigned a unique ID.
The raw data can be downloaded as an XML file 
and can be viewed online in the format 
illustrated in Figure \ref{fig:corpus}. 
Given the manageable size of the corpus, we were easily able to ensure that no duplicates 
are present. Moreover, we made sure that our corpus did not include stories
that were roughly equivalent at the level of the ASP logic form 
(e.g., we discarded stories with the same explicitly mentioned events and same 
number of customers, but with different customer names).
To make sure that the corpus is largely reproducible, we present next
the process that we used for adapting stories from the different sources 
into corpus excerpts.

\noindent
{\bf Youtube videos.} We searched \url{www.youtube.com} using the query ``ESL restaurant.'' 
We considered the first 20 results.
Many search results were videos of instructors or actors acting out a simple, stereotypical 
restaurant scenario. The videos were meant to help non-native English speakers learn 
the restaurant-related vocabulary and explain the process of dining at a restaurant with
table service. These videos were intentionally simple and thorough.
We discarded videos that were not about restaurants with table service, or 
videos that did not tell a story, and ended up with twelve videos.
A native speaker of English translated each video into a narrative that captured
not only the restaurant-related dialog, but also the actions performed by the actors.



\noindent
{\bf Google Books.} We searched Google Books using the Boolean query 
indicated by Mueller \citeyear{m07}:
{\em "the menu" AND ("the waiter brought" OR "the waiter placed" OR 
"the waiter set" OR "the waiter put" OR "the waiter poured")}. 
For each identified book, we made sure that the whole text related to the restaurant episode 
was available in Google Books. As a result, we obtained twelve book excerpts. 
From each excerpt, we removed sentences not related to what happened 
in the restaurant or conversations between characters on unrelated topics.
Finally, we transformed remaining dialogs into narratives. 

\noindent
{\bf Project Gutenberg.}
Substantially more time and effort were needed to retrieve restaurant examples
from this resource, in comparison with the previous two sources.
We downloaded 20 texts and applied the process outlined for Google Books above.
Unfortunately, many of the texts returned by the search were not useful, as they 
mentioned explicitly only two or less restaurant-related actions.
Only two stories in the corpus are obtained from Project Gutenberg.

\noindent
{\bf Mueller \citeyear{m07}.}
We included in the corpus the only sample story included in Mueller's paper (see Example \ref{ex:normal}).

\noindent
{\bf Hand-crafted.}
We considered the different types of scenarios listed in Blount \etal \citeyear{bgb15}
and determined for each of them whether a similar case could be captured by a restaurant story. 
It is important to note that the scenarios conceived by Blount \etal 
consider the perspective of a single agent thinking about its own intentions and 
making decisions about what action to execute next.
Instead, in a restaurant story, the reader learns about the actions performed by the 
several actors that are involved, but the reader is merely an observer who 
does not make decisions about actions.
We can also assume that the reader is cautious and thus does not jump to conclusions that are not 
supported by the story. For instance, when learning that 
a customer was brought a wrong dish, the cautious reader will not assume that the customer 
complained nor that he decided to eat the wrong dish, unless such information
is explicitly stated in the text. 
We believe the subset of \rest1.0 consisting of the thirteen hand-crafted excerpts
is particularly useful to the research community, 
as it covers a considerable variety of exceptional cases.


\smallskip
Inspired by Mueller's idea of automatically generating questions 
to test the understanding of restaurant narratives,
we expanded our corpus with an ASP module, also available online,
that can generate a number of queries for each excerpt. 
The module produces questions of the forms described in \cite{izbi18}:
\begin{itemize}[leftmargin=*,noitemsep,topsep=0pt]
\item $query\_yes\_no(A)$ -- Did action $A$ occur?
\item $query\_when(A)$ -- When did action $A$ occur?
\item $query\_where(P,A)$ -- Where was person $P$ when action $A$ happened?
\item $query\_who(A)$ -- Who performed action $A$?
\item $query\_who\_whom(A)$ -- Who performed action $A$ and to whom?
\item $query\_what(F,A)$ -- What was the value of fluent $F$ when action $A$ happened?
\item $query\_goal(P,A)$ -- What goal was $P$ trying to achieve when action $A$ happened?
\item $query\_intended(P,A)$ -- What was $P$'s activity/sequence of intended actions when action $A$ happened?
\end{itemize}
where $A$ is a physical action.
Queries are generated based on the \kb vocabulary and 
the entity names identified in the logic form
associated with an input text, using rules like:

$
\begin{array}{l}
n\ \{ query(yes\_no(A)) \ :\ physical\_action(A),\ \no explicit\_in\_story(A) \}\ m.\\
explicit\_in\_story(A) \ \leftarrow \ st\_hpd(A, B, S).
 
\end{array}
$

\noindent
Note that only questions about information not explicitly stated in the text are generated.
The user can set the number of questions of each type that are to be produced 
by setting the constants $n$ and $m$ to desired values.



\section{Conclusions and Future Work}
\label{sec:conclusions}

In this paper, we have extended our previous work on modeling
restaurant stories using ASP and theories of intentions.
We have shown that our approach is suitable for reasoning about stories containing
exceptions to the normal unfolding of a restaurant episode, 
which could not be processed by previous script-based approaches.
We have addressed two research questions geared towards refining
the methodology -- choice of theory of intentions and representation
of activities -- in order to increase coverage and performance. 
Additionally, we have presented a corpus of restaurant stories
that is publicly available. This will be a useful
resource for researchers in the KRR community working on stereotypical 
human activities, but also to researchers in the NLP field.

In the future, we plan to expand our corpus with new sources, new excerpts,
and new record fields to describe the excerpts (e.g., original book fragment/dialog). 
We are also working on evaluating the applicability of our methodology 
to other stereotypical activities (e.g., doctor visit)
and creating a similar story corpus for other activity domains.

\subsubsection*{Acknowledgments}
We would like to thank Zengzhi Jiang, Keya Patel, and Marcello Balduccini 
for their help in retrieving excerpts
from Google Books and Project Gutenberg.


\bibliographystyle{acmtrans}
\bibliography{TPLPrestaurant}

\end{document}